\pdfoutput=1

\documentclass[11pt]{article}

\usepackage[]{ACL2023}

\usepackage{multirow}
\usepackage{float}
\usepackage{subfig}
\usepackage{times}
\usepackage{latexsym}
\usepackage{float}
\usepackage{graphicx}
\usepackage{tabularx} 
\usepackage{ragged2e} 
\usepackage{booktabs} 
\usepackage[T1]{fontenc}

\usepackage[utf8]{inputenc}
\usepackage{cleveref} 
\usepackage{microtype}
\usepackage{stfloats}
\usepackage{hyperref} 
\usepackage{fontawesome5}
\usepackage{inconsolata}
\usepackage{url}

\usepackage{amssymb}

\title{Learning by Analogy: Diverse Questions Generation in Math Word Problem}

\author{Zihao Zhou\quad Maizhen Ning\quad Qiufeng Wang\\ \quad \textbf{Jie Yao \quad Wei Wang\quad Xiaowei Huang\quad Kaizhu Huang}\thanks{\; Duke Kunshan University, else are Liverpool University}\\
  Liverpool University \qquad
    Duke Kunshan University\\ 
    \\
 }

\begin{document}
\maketitle
\begin{abstract}
Solving math word problem (MWP) with AI techniques has recently made great progress with the success of deep neural networks (DNN), but it is far from being solved. We argue that the ability of learning by analogy is essential for an MWP solver to better understand same problems which may typically be formulated in diverse ways. However most existing works exploit the shortcut learning to train MWP solvers simply based on  samples with a single question. In lack of diverse questions, these methods merely learn shallow heuristics. 
In this paper, we make a first attempt to solve MWPs by generating diverse yet consistent questions/equations. Given a typical MWP including the scenario description, question, and equation (i.e., answer), we first generate multiple consistent equations via a group of heuristic rules. We then feed them  to a question generator together with the scenario to obtain the corresponding diverse questions, forming a new MWP with a variety of questions and equations. 
Finally we engage a data filter to remove those unreasonable MWPs, keeping the high-quality augmented ones. 
To evaluate the ability of learning by analogy for an MWP solver, we generate a new MWP dataset (called DiverseMath23K)  
with diverse questions by extending the current benchmark  Math23K.
Extensive experimental results demonstrate that our proposed method can  generate high-quality diverse questions with corresponding equations, further leading to performance improvement on DiverseMath23K.
The code and dataset is available at: 
\href{https://github.com/zhouzihao501/DiverseMWP}{https://github.com/zhouzihao501/DiverseMWP}.
\end{abstract}

\section{Introduction}
Solving Math Word Problem (MWP) aims to infer a mathematical equation and final answer from the natural language description of a math problem. Table~\ref{tab1}(a) shows one typical MWP example. In this task, the machine needs to extract relevant information from natural language texts and perform mathematical reasoning, which is challenging. 
With the boom of deep neural networks (DNN), the research of solving MWP has recently made great progress. 
For example, Seq2Seq models \citep{wang2017deep,xie2019goal,zhang2020graph} as well as pre-trained language models (PLMs) \citep{tan2021investigating,li2021seeking,liang2022mwp} have been extensively exploited to deal with MWP, and increase the prediction accuracy significantly. 
However, such models are usually in lack of the ability of learning by analogy due to the limited data size and problem diversity. Therefore, current approaches unfortunately have reached their performance bottleneck~\citep{zhang2019gap,patel2021nlp,liu2021roda,sundaram2022nlp},  showing that much remains to be done.

\begin{table}[]
\scalebox{0.55}{
\begin{tabular}{@{}l@{}}
\toprule
\textbf{(a) Original Data} \\ \midrule
\begin{tabular}[c]{@{}l@{}}\textcolor{blue}{Text: }The school makes uniforms for 40 students, known to be 15 dollars per shirt and 10\\dollars per pants. \textcolor{red}{How much did it cost to make these uniforms?} \\ \textcolor{blue}{Equation: }x = 40*(15+10)\end{tabular} \\ \midrule
\textbf{(b) Back Translation Method} \\ \midrule
\begin{tabular}[c]{@{}l@{}}\textcolor{blue}{Text: }The school produces uniforms for 40 students at \$15 per shirt and \$10 per pants. \\\textcolor{red}{How much does it cost to make these uniforms?}\\ \textcolor{blue}{Equation: }x = 40*(15+10)\end{tabular} \\ \midrule
\textbf{(c) Diverse Questions Generation} \\ \midrule
\begin{tabular}[c]{@{}l@{}}\textcolor{blue}{Scenario description: }The school makes uniforms for 40 students, known to be 15 dollars \\per shirt and 10 dollars per pants.\\ \textcolor{blue}{Question1: }How much did it cost to make a uniform?        \qquad  \textcolor{blue}{Equation1: }x = 15+10\\ \textcolor{blue}{Question2: }How much did it cost to make these shirts?        \qquad \textcolor{blue}{Equation2: }x = 40*15\\ \textcolor{blue}{Question3: }How much did it cost to make these pants?  \qquad \textcolor{blue}{Equation3: }x = 40*10\end{tabular} \\ \bottomrule
\end{tabular}
}
\caption{Examples of math word problem (MWP) generation by different methods. (a) original MWP, (b) MWP generated by back translation method~\cite{kumar2022practice}, (c) MWP with diverse questions generated by our method. The questions are highlighted by red color in the texts of (a) and (b).}
\label{tab1}
\end{table}

To alleviate this limitation, recent focus has been put on how to augment high-quality data for MWPs. Along this line, there have been some proposals~\citep {jen2021recycling,kumar2021adversarial,liu2021roda,Li2022semantic,kumar2022practice}.
Though demonstrating encouraging results, these current practices only consider word-level or sentence-level alternative expressions of the original problem, owing to the rigorous requirement in logic and numerical quantity.
As illustrated in Table~\ref{tab1}(b), the back translation augmentation method~\cite{kumar2022practice} generates less diverse data sharing very  limited semantic differences from the original counterpart. 
On the other hand, \citet{yang2022unbiased} publish a diverse MWP dataset (called UnbiasedMWP), which was collected by manual annotation with huge cost but the size is limited. 


In this paper, we make a first attempt to solve MWPs by automatically generating multiple diverse yet consistent questions (together with their corresponding equations), as illustrated in Table~\ref{tab1}(c). There are two main reasons for this augmentation strategy. (1) Training on less diverse data would lead the solver to learn shallow heuristics only, whilst deep semantics are preferred in order to better understand the problems~\cite{patel2021nlp,li2021seeking,yang2022unbiased}. Consequently, when the question is changed (i.e., \textit{Question1,2,3} in Table~\ref{tab1}(c)), the learned solver may not be able to solve MWP properly. 
(2) Our augmentation strategy could generate challenging and diverse MWPs. Training on such data would improve the ability of learning by analogy, which is essential for an MWP solver to deeply understand the problem. It is also beneficial to reduce the unreasonable case~\cite{patel2021nlp} that some current solvers still can predict the \textit{Equation} even without any question (e.g., removing the question in the text of Table~\ref{tab1}(a)).     

\begin{figure*}[t]
    \centering    
    \includegraphics[width=\textwidth]{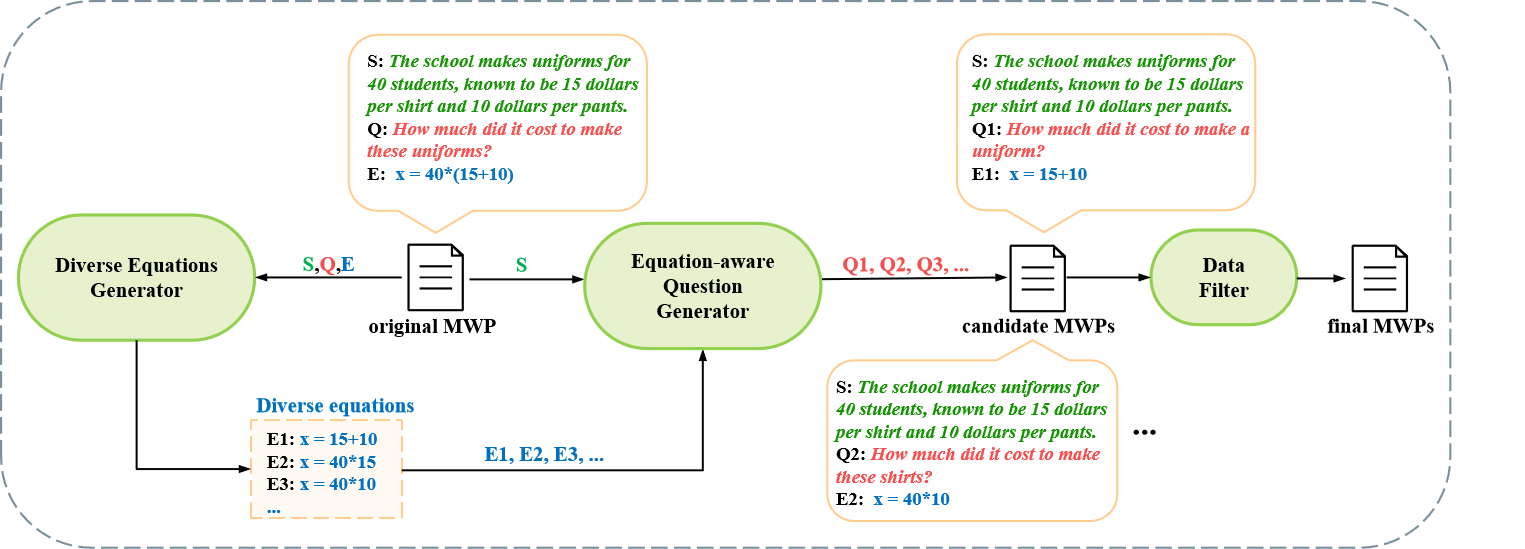}
    \caption{An overview of DQGF. Each generated equation from the \textbf{Diverse Equations Generator} and scenario description of the original MWP are fed into the trained \textbf{Equation-aware Question Generator} to generate corresponding questions. In this way, we will obtain diverse questions with their equations and form a new MWP. Finally, the candidate MWPs are further filtered using the \textbf{Data Filter.}}
    \label{fig1}
    \end{figure*}

Motivated by these findings, we propose a \textbf{D}iverse \textbf{Q}uestions \textbf{G}eneration \textbf{F}ramework (\textbf{DQGF}) to generate high-quality and diverse questions with their corresponding equations for a given MWP. Our DQGF consists of three components as shown in Figure~\ref{fig1}. (1) \textbf{Diverse Equations Generator:} It generates diverse and meaningful equations from the original MWP based on two generation strategies. Specifically, we propose a sub-equation based strategy that extracts sub-equations from the original equation, and a unit based strategy that generates equations according the units (e.g., "dollars" in Table~\ref{tab1}) in the scenario description. (2) \textbf{Equation-aware Question Generator:} Given a scenario description and generated equation, it generates a corresponding question. For example, given the \textit{Scenario description} and \textit{Equation1} in Table~\ref{tab1}(c), it can generate \textit{Question1}. In details, we utilize two encoders to extract the information of scenario description and equation respectively, and design an interaction mechanism which exploits numbers as a bridge to fuse the information of both encoders. (3) \textbf{Data Filter}: A large-scale MWP pre-trained language model~\cite{liang2022mwp} is leveraged to filter unreasonable data. As such, we can generate many high-quality and diverse MWP samples. 

Extensive experiments on the existing dataset UnbiasedMWP~\cite{yang2022unbiased} show that our proposed DQGF could generate high-quality diverse questions with corresponding equations, thus increasing the accuracy of the MWP solver. To further verify the effectiveness of the DQGF, we produce a new dataset (called DiverseMath23K) with diverse questions from the current benchmark dataset Math23K~\cite{wang2017deep}. We also propose a new Group-accuracy metric on all questions of a problem. Experimental results show that DQGF can effectively improve the overall performance of the solver on DiverseMath23K, demonstrating  its ability of learning by analogy.
In summary, our contributions are as follows:
\begin{itemize}
\item{}We propose a novel diverse questions generation framework (DQGF) to automatically generate diverse questions with their corresponding equations for a given MWP. To the best of our knowledge, this is the first effort to generate such data in MWP. 
\item{}We propose a Diverse Equations Generator, consisting of sub-equations based  and unit based strategy to generate diverse and meaningful equations from the original MWP.
\item{}We propose an Equation-aware Question Generator to generate a question from the given scenario and equation. It consists of two encoders to encode scenario and equation respectively where an interaction mechanism is developed to fuse the information.
\item{}We produce a new MWP dataset (called DiverseMath23K) with diverse questions by extending the current benchmark Math23K.
\item{}Experimental results demonstrate that DQGF could generate high-quality diverse questions and improve effectively the overall performance of the MWP solver on both UnbiasedMWP and DiverseMath23K.
\end{itemize}

\section{Related Work}
\paragraph{Data Augmentation:} Data augmentation has
been widely used in various NLP tasks \cite{feng2021survey}, but there are few works for MWP. Recently, some MWP data augmentation methods have been proposed. For example, 
\citet{kumar2021adversarial} reorder the problem description like moving the question at the start. Furthermore, they paraphrase sentences by a paraphrasing model and preserve the entities of the original sentence to keep the theme unchanged. \citet{kumar2022practice} further propose back translation, synonym replacement, and named-entity replacement to augment data. \citet{Li2022semantic} and \citet{liu2021roda} transform the declarative sentence into the question sentence and reverse the operation of expression to generate MWPs.  These methods effectively improve the performance of MWP solvers. But most of them are rule-based and augment data with limited semantic differences from the original data.
\paragraph{MWP Solver:} Recent proposals intend to solve the problem by using sequence or tree generation models. \citet{wang2017deep} present a sequence-to-sequence (seq2seq) approach to generate the mathematical equation. \citet{xie2019goal} propose a goal-driven tree-structured (GTS) model to generate the equation tree. This sequence-to-tree approach significantly improves the performance over the traditional seq2seq approaches. \citet{zhang2020graph} adopt a graph-to-tree
approach to model the quality relations using graph convolutional networks (GCN). Applying pre-trained language models such as BERT \cite{2019bert} was shown to benefit the tree expression generation substantially. Prior study \citep {patel2021nlp} indicates that existing MWP solvers rely on shallow heuristics to generate equations. As such, they could not solve different questions of the same MWP well and even ignore the question. Our DQGF effectively helps the solver overcome these issues. 
\paragraph{MWP Generation:} MWP generation approaches can be divided into three categories: template-based approaches, rewriting-based approaches, and neural network-based approaches. Template-based approaches usually follow a similar two-stage process: they first generalize an existing problem into a template or a skeleton and then generate the MWP sentences from the templates~\cite{williams2011generating,polozov2015personalized}. Rewriting-based approaches target the MWP generation problem by editing existing human-written MWP sentences to change their theme but  the underlying story \cite{koncel2016theme,moon2019illmatics}. Recent attempts have been focused on exploiting neural network-based approaches that generate MWPs from equations and topics in an end-to-end manner \cite{liyanage2020multi,liu2021mathematical,wang2021math}. Unlike these generation methods, our equation-aware question generator focuses on generating questions that are in line with the given scenario and match the given equation. Recently, \citet{shridhar2022automatic} have also proposed a generation model to implement this function, but main differences exist: (1) Their work focuses on generating goal-driven sub-questions without equations, which is used in prompt learning instead of a general data augmentation tool. (2) While their generator directly concatenates the scenario and equation text sequence to encode and fuse their information, the structure of equation is much different from the scenario texts. We propose two different encoders where an interaction mechanism is designed to leverage numbers as a bridge to  fuse the information. 

\paragraph{MWP Dataset:} Several datasets are proposed to evaluate the model’s numerical reasoning ability \cite{koncel2016mawps,wang2017deep,amini2019mathqa,miao2020diverse}. They only provide
 a single question to each scenario. Therefore, training and evaluating on such setting will lead  that the solvers  rely on shallow heuristics to generate equations \cite{patel2021nlp}. To mitigate this learning bias, \citet{yang2022unbiased} propose a diverse MWP dataset (called UnbiasedMWP). However, manually collecting high-quality datasets is usually labor-intensive and time-consuming in practice. In contrast, our DQGF could automatically generate such diverse data. In this paper, we will use UnbiasedMWP to train equation-aware question generator and evaluate the whole DQGF. Besides, we also propose a diverse MWP dataset DiverseMath23k to evaluate the MWP solver.

\section{Methodology}
Figure~\ref{fig1} shows the overview of the proposed \textbf{D}iverse \textbf{Q}uestions \textbf{G}eneration \textbf{F}ramework (\textbf{DQGF}). We firstly put the original MWP into the Diverse Equations Generator to generate diverse equations, then the generated equation and scenario description of the original MWP are fed into the trained equation-aware question generator to produce corresponding questions. In this way, we will obtain diverse questions with their equations, forming new candidate MWPs. Finally, these candidate MWPs are further filtered by the data filter. In what follows, we will introduce Diverse Equations Generator, Equation-aware Question Generator, and Data Filter respectively in Section~\ref{sec32}, Section~\ref{sec31}, and Section~\ref{sec33}.
\subsection{Diverse Equations Generator}\label{sec32}
Diverse equations generator aims to generate diverse equations from the original MWP. Our principle is to generate as many as possible logical equations. Motivated by this, we propose two equation generation strategies: sub-equation based  and unit based strategy.
\paragraph{Sub-equation Based}
The equation of the original MWP usually includes some sub-equations, which represent the necessary steps to solve the problem~\cite{2021Training}. For instance, in Table~\ref{tab1}(c), "15+10" is a sub-equation of the original equation, describing a uniform's price. Therefore, we extract these sub-equations from the original equation, which are very high-quality and diverse.
\paragraph{Unit Based}
There are some physical relations between the numbers in an MWP. We could identify these relations, and then combine numbers with operators to get a new equation. 
Motivated by this, we propose to search the relations of numbers based on their units. 
Every number in MWPs has its unit. For example in Table~\ref{tab1}, "40" has the unit "students" and "15" has the unit "dollars". We combine them in two situations. (1) Same unit: Two numbers with same unit always represent the same object. We combine them with the operator "+" to generate equations representing the totality questions like "what is the total of A and B". Besides, we combine them with "-" and "/"  which represent the comparison questions like "how much more A than B" and "how many times A than B", respectively. (2) Different units: Two numbers with different units in a MWP always represent two objects that have subordinate relations. Therefore, we combine them with "*". This strategy will generate diverse equations, though it probably brings some unreasonable equations further generating noisy MWPs. Such noisy MWPs will be filtered by the final data filter.

To be noted, both sub-equation based and unit based strategies rely on heuristic rules. Therefore, we do not need to train our diverse equations generator.

\subsection{Equation-aware Question Generator}\label{sec31}
\begin{figure}[t]
    \centering    
    \includegraphics[width=\linewidth]{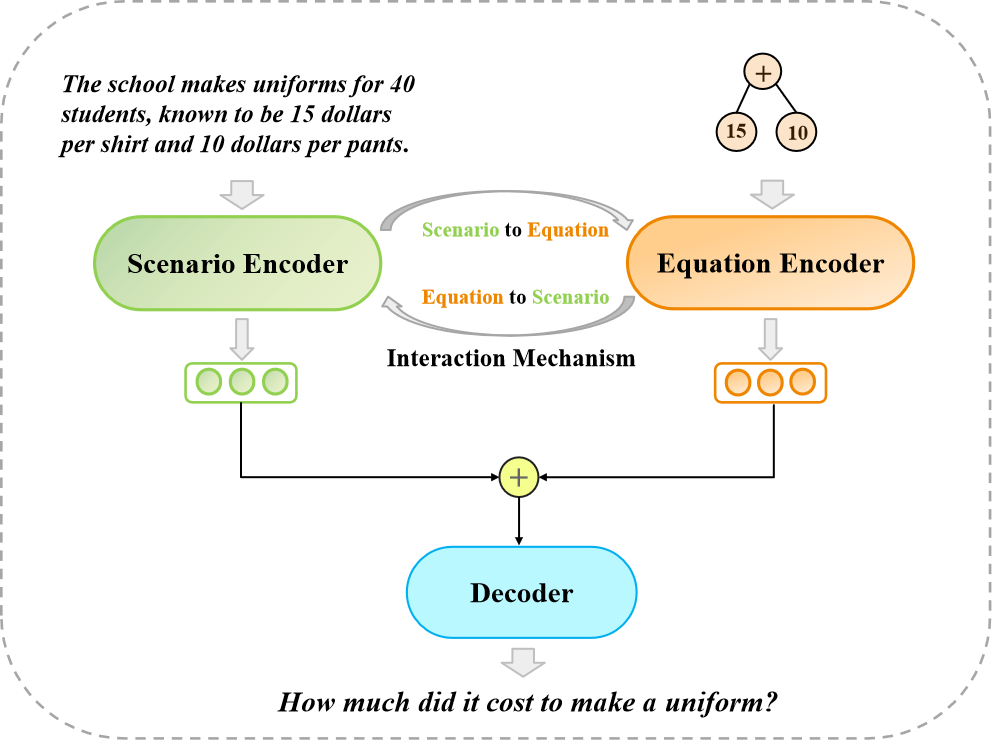}
    \caption{Equation-aware Question Generator}
    \label{fig2}
\end{figure}
General question generation in the Question-Answering area aims to generate a question from a given passage and a specified answer~\cite{sun2018answer,kim2019improving,li2019improving}.
By regarding the scenario description and equation as passage and answer respectively, we can formulate our task as a general question generation problem. Based on this, we propose an equation-aware question generator under a general encoder-decoder framework as shown in Figure~\ref{fig2}. Specifically, we propose two different encoders to encode the information from scenario and equation respectively, and an interaction mechanism to fuse their information further. For convenience, we form a MWP as $\left(S,Q,E\right)$ , where $S$, $Q$ and $E$ represent the scenario, question and their solution equation respectively. 


\paragraph{Scenario Encoder}
We adopt a pre-trained language model BERT \cite{2019bert} 
as our scenario encoder. The unsupervised pre-training on large corpora makes the model capture linguistic knowledge, which provides rich textual representations. We represent the scenario $S$ as a sequence of \textit{T} tokens: $S = \left[ s_1,s_2,...,s_T\right]$, and 
formulate the encoding process as \begin{equation}\left[ h_1^s,h_2^s,...,h_T^s \right] = BERT\left( \left[ s_1,s_2,...,s_T\right]\right),\label{equation1}\end{equation}
where $h_i^s$ represents the embedding of token $s_i$ from the encoder. Finally, the representation of scenario can be written as $H^s$: \begin{equation}
    H^s = \left[ h_1^s,h_2^s,...,h_T^s \right].\end{equation}
    
\paragraph{Equation Encoder}
The sequence form cannot model the structure of the equation well \cite{xie2019goal}. Hence we transform it into an equation tree which is then encoded by a TreeLSTM \cite{tai2015improved}. The equation is transformed into a binary tree representation as proposed in \cite{xie2019goal} and sequentialized as their pre-order traversal. Thus the equation can be represented as $E = \left[e_1,e_2,...,e_n\right]$, where \textit{n} is the length of the pre-order equation and a node $e_i$ represents a number or operator (+,-,*,/). In details, we firstly adopt a BERT to encode each node: \begin{equation}
    x_i = BERT\left(e_i\right).\label{equation3}\end{equation}
Then, we encode the equation tree by a TreeLSTM: \begin{equation}
    h_i^e = TreeLSTM\left(x_i, \sum_{k\in C\left(i\right)}h_k^e\right),  \label{equation4} \end{equation}
where $C\left(i\right)$ represents the index set of child nodes of $e_i$. 
Finally, the representation of equation can be written as $H^e$: \begin{equation}
    H^e = \left[ h_1^e,h_2^e,...,h_n ^e \right].\end{equation}

\paragraph{Interaction Mechanism}
In order to generate a question based on both scenario and equation, the interaction between them is crucial. Inspired by iterative deep learning \cite{he2019deepotsu,schick2020exploiting}, we propose an interaction mechanism which uses numbers as bridge to fuse the information of both scenario and equation. It consists of the following two processes.

\textit{Scenario to Equation}: After BERT encodes the whole scenario text, each token's embedding has the scenario's context information. For a number appearing in both scenario and equation, we replace its embedding in Equation~(\ref{equation3}) with its embedding in Equation~(\ref{equation1}). In this way, the scenario's context information is brought into the equation.

\textit{Equation to Scenario}: After bringing the information of the scenario to the equation and encoding the equation tree, we put the embedding of the number in the equation back into the scenario representation. In detail, we replace its embedding in Equation~(\ref{equation1}) with its embedding in Equation~(\ref{equation4}).

\paragraph{Decoder}
We adopt the pre-trained language model BertGeneraiton \cite{rothe2020leveraging} as our decoder. Representing a question $Q$ as a sequence of \textit{m} tokens: $Q = \left[ q_1,q_2,...,q_m\right]$, the token $q_i$ is generated as\begin{equation}
    q_i = BertGeneration\left(\left[H,q_{i-1} \right]\right),\end{equation}
where $H$ is the final representation of the scenario and equation by the concatenating the $H_s$ and $H_e$ as \begin{equation}
    H = \left[H_s,H_e \right].\end{equation}

To be noted, all of these pre-trained models in both encoders and decoders will be fine-tuned in the MWP dataset.

\subsection{Data Filter}\label{sec33}
Filtering out detrimental augmented data can improve the quality of data as well as the downstream performance~\cite{le2020adversarial}. However, it will take a great cost to do it by the human filtering due to the large-size of our augmented data. Therefore, we utilize an existing powerful MWP solver as an expert model to judge whether the predicted answer is same as the ground-truth~\cite{axelrod2011domain,xie2021target}. Inspired by \citet{ou2022counterfactual}, we leverage a large-scale MWP pre-trained language model MWP-BERT \cite{liang2022mwp} as our expert model, utilizing its powerful generalization ability. 

Considering our generated MWPs have many new diverse questions, it is difficult for an existing solver to predict the answer accurately, resulting in many false filtering cases. To increase the recall on the generated samples, we apply beam-search strategy on the expert model to select top $k$ predicted equations (We set $k=5$ in our experiments). Since the final answer can be from different solutions~\cite{yang2022unbiased}, we compare the answer calculated by equations instead of comparing equations directly. The augmented MWPs will pass our final filter if its final answer is equal to one answer from the selected top $k$ equations predicted by the expert model.



\section{Experiments}

\subsection{Dataset and experimental setting} 
\paragraph{Dataset} We conduct experiments on an existing diverse questions dataset: UnbiasedMWP~\cite{yang2022unbiased}, which is split into 2,507, 200, 200 MWP groups for training, validation, and testing, respectively. Each group contains one original MWP and additional 1 to 8 diverse questions and equations with the same scenario. In total, it has 8,895, 684, 685 MWPs for training, validation, and testing, respectively. In this paper, we train our Equation-aware Question Generator and evaluate the whole DQGF on it.

\paragraph{Evaluation Metrics} 
For the whole DQGF, we apply the accuracy of a MWP solver to evaluate the quality of generated data. Without loss of generality, we choose GTS~\cite{xie2019goal} with BERTEncoder \cite{2019bert} as the MWP solver. Furthermore, we also propose a metric of Group-Accuracy to consider the prediction accuracy on all diverse questions in a MWP. For example, in Table~\ref{tab1}(c), the normal accuracy simply 
regards it as three samples by evaluation of each question separately, while our Group-Accuracy considers this as only one sample and if all three equations are predicted correctly then the prediction is correct. 
Comparing to the common accuracy, the proposed Group-Accuracy can evaluate an solver whether truly understanding an MWP with the ability of learning by analogy. 
For the equation-aware question generator, we report BLEU \cite{papineni2002bleu}, ROUGE-1, ROUGE-2, and ROUGE-L\cite{lin2004rouge} which are based on exact word overlapping. BERT F1 score \cite{zhang2019bertscore} is also used, which is based on DeBERTa \cite{he2020deberta}.



\subsection{Experimental Results} 
\begin{table}[]
\scalebox{0.91}{
\begin{tabular}{lll}
\hline
\textbf{Data}                         & \textbf{Accuracy} & \textbf{Group-Accuracy} \\ \hline
{Unbiased-source} & 34.9              & 29.5                    \\
\textbf{Unbiased-DQGF}                & \textbf{62.7}     & \textbf{42.0}           \\ \hline
Unbiased-GT               & 78.4     & 64.0         \\ \hline
\end{tabular}
}
\caption{Comparison of the accuracy (\%) of solver training on different data: Unbiased-source means the original MWPs of each group in UnbiasedMWP, Unbiased-DQGF means generated MWPs with diverse questions by our DQGF, Unbiased-GT means MWPs with the annotated diverse questions and equations, indicating the up-bounded performance of our DQGF.}
\label{tab2}
\end{table}

We evaluate the quality of generated data by the results of a common MWP solver on both accuracy and group-accuracy. In details, we train the MWP solver on three different data: the original data of each group in the UnbiasedMWP (called Unbiased-source), our generated MWPs data from the UnbiasedMWP (called Unbiased-DQGF), and ground-truth MWPs in the UnbiasedMWP (called Unbiased-GT). Notably, the Unbiased-source only has MWPs with single question, while the latter two have MWPs with diverse questions. Since the Unbiased-GT directly uses the annotated diverse questions, its performance can be regarded as the up-bounded of the generation method. The results are shown in Table~\ref{tab2}.

As shown in Table~\ref{tab2},  
we can see that training on the data augmented by DQGF can significantly improve the accuracy of solver from 34.9\% to 62.7\%. It indicates that DQGF can generate high quality MWP samples, which are useful for the training of a solver. In addition, the group-accuracy is also increased largely from 29.5\% to 42\%, even higher than the common accuracy (34.9\%) of Unbiased-source, showing that our method can generate MWP samples with valid diverse questions to help the solver better understand the problem by capturing the ability of learning by analogy. 
Comparing the Unbiased-DQGF and Unbiased-GT, we can see that there is still a gap between our method and the manual labelling data. Manual annotation method can produce more diverse and completely correct data, which leads to the better performance. 

\subsection{Fine-grained Analysis}
In this section, we will show the performance of the three components in our DQGF individually.  

\begin{table}[]
\centering
\scalebox{0.92}{
\begin{tabular}{ll}
\hline
\textbf{Strategy}   & \textbf{Accuracy} \\ \hline
\textbf{All}       & \textbf{62.7}     \\
(w/o)Sub-equations  & 58.5              \\
(w/o)Same unit      & 47.3              \\
(w/o)Different units & 60.4              \\ \hline
\end{tabular}
}
\caption{Comparison of different equations generation strategies.}
\label{tab4}
\end{table}

\paragraph{Diverse Equations Generator} Table~\ref{tab4} shows the comparison results among different equations generation strategies. As observed, each strategy can generate high quality and meaningful diverse equations. Concretely, the same unit based generation strategy brings the most benefit to DQGF because such strategy can generate a lot of meaningful but less noisy equations. 
The sub-equations based strategy and different units based strategy can also effectively generate meaningful equations, but with little improvement to the solver. There are two reasons: 1) The sub-equations based strategy can not generate enough equations since the sub-equations in the original equation are limited; and 2) The different units based strategy generates meaningful equations while bringing many noisy equations, which are thus hard to be filtered completely.

\begin{table}[]
\centering
\scalebox{0.62}{
\begin{tabular}{llllll}
\hline
\textbf{Methods} & \textbf{BLEU} & \textbf{BERT F1} & \textbf{ROUGE-1} & \textbf{ROUGE-2} & \textbf{ROUGE-L} \\ \hline
Baseline         & 52.3 & 87.4 & 77.2 & 59.4 & 70.6 \\
EQG(w/o)IM & 54.2 & 87.9 & 78.4 & 61.4 & 72.0 \\
\textbf{EQG}    & \textbf{60.5} & \textbf{89.7} & \textbf{81.4} & \textbf{66.7} & \textbf{77.4} \\ \hline
\end{tabular}
}
\caption{Comparison of the different questions generator models. The baseline directly concatenates the scenario and equation text sequence. EQG means the Equation-aware Question Generator, while EQG(w/o)IM means removing Interaction Mechanism.}
\label{tab3}
\end{table}

\paragraph{Equation-aware Question Generator} 
We compare one baseline method that directly concatenates the scenario and equation text sequence \cite{shridhar2022automatic} and utilizes BERT \cite{2019bert} as encoder, and BertGeneration \cite{rothe2020leveraging} as decoder.
Table~\ref{tab3} reports the comparison of the different questions generator models. 
We can see that EQG(w/o)IM improves the performance of baseline method. It indicates that the scenario encoder and equation encoder can better encode the structure of scenario and equation respectively than directly encoding their concatenated sequence. 
By integrating the interaction mechanism (IM), we can observe that it leads to a great improvement, achieving 
the best performance on every metric, which demonstrates that our interaction mechanism can fuse the information of scenario and equation well. 
Specifically, the BLEU score is 60.5\% which is not high; this is however explainable as it is a metric about text overlap. As observed, though semantically identical, some of our generated data is less overlap with the ground truth. This can also be reflected by its higher BERT F1 score which measures the semantic similarity.

\begin{figure}[t]
    \centering
    \includegraphics[width=0.8\linewidth]{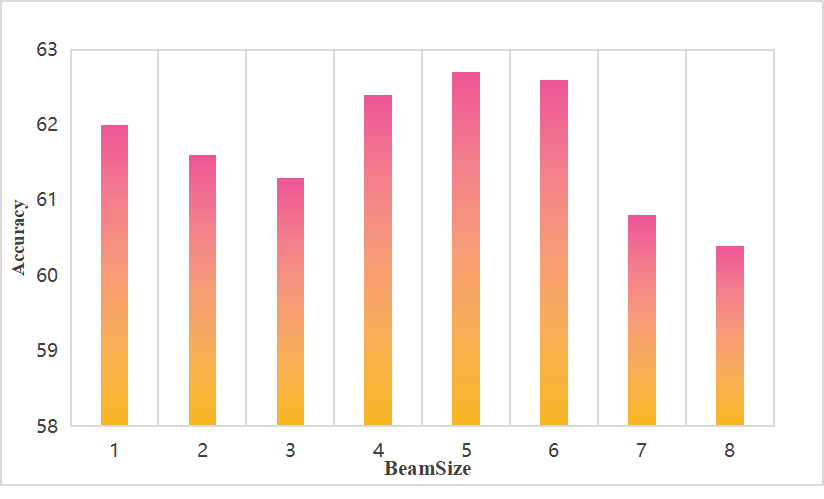}
    \caption{Different beamsize \textit{k} of expert model in Filter.}
    \label{fig3}
\end{figure}
    
\paragraph{Data Filter} We examine the effect of beamsize \textit{k} of the filter in DQGF, which is shown in Figure~\ref{fig3}. The experimental results show that DQGF can obtain the best performance when \textit{k} is 5. DQFG can achieve good performance when \textit{k} is between 4 and 6, since this appears to be a suitable interval in that a lot of correct candidates can pass the filter. When \textit{k} is between 1 and 3, filtering is still accurate but some correct data are filtered out. Therefore this interval can achieve competitive but not the best performance. When \textit{k} is between 7 and 8, the filtering is inaccurate. It causes that some noisy data pass the filter and impacts the final data quality.

\subsection{New MWP dataset DiverseMath23K}

We apply our trained DQGF model on Math23k to create a new MWP dataset (called
DiverseMath23K) with diverse questions, which contains 38,320, 1,255, 1,728 MWPs for training, validation, and testing respectively. 

To ensure the quality of DiverseMath23k, we manually check generated MWPs, which is much easier and more efficient than  complete human annotation.
For the validation and test set, to make the evaluation rational, we rigorously check and correct each sample by ourselves. For the training set, we randomly check parts of samples and find that our generated MWPs are also meaningful and credible. 
The final dataset is available at \href{https://github.com/zhouzihao501/DiverseMWP}{https://github.com/zhouzihao501/DiverseMWP}.


\begin{table}[]
\scalebox{0.8}{
\begin{tabular}{llll}
\hline
\textbf{Data} & \textbf{Accuracy} & \textbf{Group-Accuracy} & \textbf{Deq-Accuracy} \\ \hline
Ori           & 63.6              & 56.9                    & 69.4                  \\
Diverse          & \textbf{68.4}     & \textbf{60.2}           & \textbf{48.1}         \\ \hline
\end{tabular}
}
\caption{Performance of solvers training on different data. Ori and DQGF means the original Math23k and DiverseMath23k, respectively.}
\label{tab7}
\end{table}

\begin{table}[]
\scalebox{0.75}{
\begin{tabular}{l}
\hline
\begin{tabular}[c]{@{}l@{}}\textbf{Qriginal MWP:}\\ The candy in the mall costs 14.60 dollars per box and
cookies\\ cost 29.80 dollars per box. Uncle Li wants to buy 4 boxes of \\candy and 2 boxes of cookies. Please calculate how much money\\ Uncle Li needs to bring?\\ Equation: x=(14.6*4)+(29.8*2)\end{tabular} \\ \hline
\textbf{Generated Data:} \\ \hline
\begin{tabular}[c]{@{}l@{}}\textcolor{red}{Question: }how many dollars it will cost to buy the cookies?\\ \textcolor{blue}{Equation: }x=29.8*2 \\ Equation type: sub-equation, different units\end{tabular} \\ \hline
\begin{tabular}[c]{@{}l@{}}\textcolor{red}{Question: }how much more expensive each box of cookies \\is than each box of candy?\\ \textcolor{blue}{Equation: }x=29.8-14.6\\ Equation type: same unit\end{tabular} \\ \hline
\begin{tabular}[c]{@{}l@{}}\textcolor{red}{Question: }how many times the price of each box of candy is \\the price of each box of cookies?\\\textcolor{blue}{Equation: } x=14.6/29.8\\ Equation type: same unit\end{tabular} \\ \hline
\end{tabular}}
\caption{Generated \textcolor{red}{diverse questions} with \textcolor{blue}{equations} and their corresponding equation types}
\label{tab5}
\end{table}

\paragraph{Results} 
We compare the performance of the solver training on original Math23k and DiverseMath23k. In addition to the accuracy and Group-Accuracy, we report the Deq-Accuracy~\cite{patel2021nlp}, which is a metric measuring the question sensitivity. The lower the Deq-Accuracy, the better the question sensitivity. 
Concretely, it measures the accuracy that the solver predicts the answer of a MWP by \textbf{de}leting \textbf{q}uestions (i.e., only input scenario). A better solver should have higher question sensitivity, thus a lower Deq-Accuracy is expected.

The results are shown in Table~\ref{tab7}. We can see that the accuracy can be improved from 63.6\% to 68.4\%, and Group-Accuracy is boosted from 56.9\% to 60.2\%. These results indicate that DiverseMath23k can enable the model to better understand MWPs and improve its ability to solve different questions in the same scenario, even our training set possibly cantains many noisy samples. Additionally, it is noted that our method can significantly reduce the Deq-accuracy from 69.4\% to 48.1\%. It indicates that DiverseMath23k effectively improves the question sensitivity of the solver.

\subsection{Case Study}

\begin{table}[]
\centering
\scalebox{0.69}{
\begin{tabular}{llll}
\hline
\begin{tabular}[c]{@{}l@{}}Scenario: A factory produce 3000 parts, 750\\ in the first 6 days and the rest in 15 days\end{tabular} &
 &
  \textbf{Ori} &
  \textbf{Diverse} \\ \hline
\begin{tabular}[c]{@{}l@{}}Question1: How many will be produced on\\ average per day in the future? (\textbf{original question})\\ Equation1: x=(3000-750)/15\end{tabular} &
   &
  \textcolor{green}{True} &
  \textcolor{green}{True} \\ \hline
\begin{tabular}[c]{@{}l@{}}Question2: How many more will be produced?\\ Equation2: x=3000-750\end{tabular} &
   &
  \textcolor{red}{False} &
  \textcolor{green}{True} \\ \hline
\begin{tabular}[c]{@{}l@{}}Question3: What is the average number of parts\\ produced per day for the first 6 days?\\ Equation3: x=750/6\end{tabular} &
   &
  \textcolor{red}{False} &
  \textcolor{green}{True} \\ \hline
\end{tabular}
}
\caption{Prediction results of solvers training on different data: Ori means original Math23k, and Diverse means DiverseMath23k.}
\label{tab6}
\end{table}

\paragraph{Generated Data Analysis}Table~\ref{tab5} shows some real cases  generated by our DQGF. We can see that our Diverse Equation Generator generate multiple meaningful equations. Moreover, the same unit based strategy can generate the most. After getting the diverse equations, our Equation-aware Question Generator successfully generates corresponding questions that match the scenario and equations. In particular, Equation-aware Question Generator works well in relating objects with their corresponding numbers. Therefore the appearance order of objects in questions are not reversed. Finally, these correct MWPs can successfully pass the data filter. More generated samples are shown in Appendix~\ref{sec:appendix}.
\paragraph{Prediction Results Analysis} Table~\ref{tab6} reports the prediction result of solvers trained on different data. The solver trained on the original Math23k can correctly solve Question1, which has a similar MWP in training. However, it cannot solve Question2, which is simpler than Question1. Moreover, it cannot solve other questions like Question3. It indicates that the solver merely learns shallow heuristics but failing to understand the MWP. When trained on DiverseMath23k, the solver would gain the ability of learning by analogy, i.e., the solver could solve different questions even if the question is changed (see Question2, and Question3).

\section{Conclusion and Future Work}
In this paper, we explore the ability of learning by analogy for MWP solvers. To do this, we propose a diverse questions generation framework (DQGF) to automatically generate diverse questions with their corresponding equations for a give MWP, which consists of Diverse equations Generator, Equation-aware Question Generator and Data Filter. 
Based on the trained DQGF, we further produce a new MWP dataset (DiverseMath23K) with diverse questions. Experimental results demonstrate that DQGF could generate high-quality diverse questions and improve effectively the overall performance of the MWP solver.

In the future, we will focus on optimizing the model in the solver to improve its ability of learning by analogy and increase the group accuracy on the MWPs with diverse questions.

\section*{Limitations}
Our DQGF still exists some limitations. While our generated data improves performance in diverse questions settings, there is still some noise in the generated data that affects the performance of original single question. In the following, we will give the limitations of our DQGF on its three components.

The diversity of the question depends on the diversity of the equations. Our equation generator is based on heuristic rules, resulting that the generated equations are very simple. In the future, we will try a model based equations generator to generate more diverse equations. 
In the question generator, it can only recognise equations with the operator "+-*/" due to the limited operator set in our training dataset UnbiasedMWP. 
In the future we will expand the operators so that the generation model can recognise more operators and be more universal. Filtering strategy is also important. Using the answers of expert model as a criterion for evaluation still exists bias and leads to the noisy data. In fact, we have tried to generate more diverse equations but all are filtered by the current data filter. We will look for better filtering strategies in the future.

\section*{Acknowledgements}
This research was funded by National Natural Science Foundation of China (NSFC) no.62276258, Jiangsu Science and Technology Programme (Natural Science Foundation of Jiangsu Province) no. BE2020006-4, Xi’an Jiaotong-Liverpool University's Key Program Special Fund no. KSF-T-06, European Union’s Horizon 2020 research and innovation programme no. 956123, and UK EPSRC under projects [EP/T026995/1].

\bibliography{ref}

\begin{thebibliography}{44}
\expandafter\ifx\csname natexlab\endcsname\relax\def\natexlab#1{#1}\fi

\bibitem[{Amini et~al.(2019)Amini, Gabriel, Lin, Koncel-Kedziorski, Choi, and
  Hajishirzi}]{amini2019mathqa}
Aida Amini, Saadia Gabriel, Shanchuan Lin, Rik Koncel-Kedziorski, Yejin Choi,
  and Hannaneh Hajishirzi. 2019.
\newblock \href {https://doi.org/10.18653/v1/N19-1245} {{M}ath{QA}: Towards
  interpretable math word problem solving with operation-based formalisms}.
\newblock In \emph{Proceedings of the 2019 Conference of the North {A}merican
  Chapter of the Association for Computational Linguistics: Human Language
  Technologies, Volume 1 (Long and Short Papers)}, pages 2357--2367,
  Minneapolis, Minnesota. Association for Computational Linguistics.

\bibitem[{Axelrod et~al.(2011)Axelrod, He, and Gao}]{axelrod2011domain}
Amittai Axelrod, Xiaodong He, and Jianfeng Gao. 2011.
\newblock Domain adaptation via pseudo in-domain data selection.
\newblock In \emph{Proceedings of the 2011 conference on empirical methods in
  natural language processing}, pages 355--362.

\bibitem[{Cobbe et~al.(2021)Cobbe, Kosaraju, Bavarian, Hilton, Nakano, Hesse,
  and Schulman}]{2021Training}
K.~Cobbe, V.~Kosaraju, M.~Bavarian, J.~Hilton, R.~Nakano, C.~Hesse, and
  J.~Schulman. 2021.
\newblock Training verifiers to solve math word problems.

\bibitem[{Devlin et~al.(2019)Devlin, Chang, Lee, and Toutanova}]{2019bert}
Jacob Devlin, Ming{-}Wei Chang, Kenton Lee, and Kristina Toutanova. 2019.
\newblock \href {https://doi.org/10.18653/v1/n19-1423} {{BERT:} pre-training of
  deep bidirectional transformers for language understanding}.
\newblock In \emph{NAACL}.

\bibitem[{Feng et~al.(2021)Feng, Gangal, Wei, Chandar, Vosoughi, Mitamura, and
  Hovy}]{feng2021survey}
Steven~Y. Feng, Varun Gangal, Jason Wei, Sarath Chandar, Soroush Vosoughi,
  Teruko Mitamura, and Eduard~H. Hovy. 2021.
\newblock \href {https://doi.org/10.18653/v1/2021.findings-acl.84} {A survey of
  data augmentation approaches for {NLP}}.
\newblock In \emph{Findings of the Association for Computational Linguistics:
  {ACL/IJCNLP} 2021, Online Event, August 1-6, 2021}, volume {ACL/IJCNLP} 2021
  of \emph{Findings of {ACL}}, pages 968--988. Association for Computational
  Linguistics.

\bibitem[{He et~al.(2021)He, Liu, Gao, and Chen}]{he2020deberta}
Pengcheng He, Xiaodong Liu, Jianfeng Gao, and Weizhu Chen. 2021.
\newblock \href {https://openreview.net/forum?id=XPZIaotutsD} {Deberta:
  decoding-enhanced bert with disentangled attention}.
\newblock In \emph{9th International Conference on Learning Representations,
  {ICLR} 2021, Virtual Event, Austria, May 3-7, 2021}. OpenReview.net.

\bibitem[{He and Schomaker(2019)}]{he2019deepotsu}
Sheng He and Lambert Schomaker. 2019.
\newblock Deepotsu: Document enhancement and binarization using iterative deep
  learning.
\newblock \emph{Pattern recognition}, 91:379--390.

\bibitem[{Jen et~al.(2021)Jen, Huang, and Chen}]{jen2021recycling}
Tien-Yi Jen, Hen-Hsen Huang, and Hsin-Hsi Chen. 2021.
\newblock Recycling numeracy data augmentation with symbolic verification for
  math word problem solving.
\newblock In \emph{IEEE/WIC/ACM International Conference on Web Intelligence
  and Intelligent Agent Technology}, pages 653--657.

\bibitem[{Kim et~al.(2019)Kim, Lee, Shin, and Jung}]{kim2019improving}
Yanghoon Kim, Hwanhee Lee, Joongbo Shin, and Kyomin Jung. 2019.
\newblock Improving neural question generation using answer separation.
\newblock In \emph{Proceedings of the AAAI conference on artificial
  intelligence}, volume~33, pages 6602--6609.

\bibitem[{Koncel{-}Kedziorski et~al.(2016)Koncel{-}Kedziorski, Konstas,
  Zettlemoyer, and Hajishirzi}]{koncel2016theme}
Rik Koncel{-}Kedziorski, Ioannis Konstas, Luke Zettlemoyer, and Hannaneh
  Hajishirzi. 2016.
\newblock \href {https://doi.org/10.18653/v1/d16-1168} {A theme-rewriting
  approach for generating algebra word problems}.
\newblock In \emph{Proceedings of the 2016 Conference on Empirical Methods in
  Natural Language Processing, {EMNLP} 2016, Austin, Texas, USA, November 1-4,
  2016}, pages 1617--1628. The Association for Computational Linguistics.

\bibitem[{Koncel-Kedziorski et~al.(2016)Koncel-Kedziorski, Roy, Amini, Kushman,
  and Hajishirzi}]{koncel2016mawps}
Rik Koncel-Kedziorski, Subhro Roy, Aida Amini, Nate Kushman, and Hannaneh
  Hajishirzi. 2016.
\newblock Mawps: A math word problem repository.
\newblock In \emph{Proceedings of the 2016 Conference of the North American
  Chapter of the Association for Computational Linguistics: Human Language
  Technologies}, pages 1152--1157.

\bibitem[{Kumar et~al.(2021)Kumar, Maheshwary, and Pudi}]{kumar2021adversarial}
Vivek Kumar, Rishabh Maheshwary, and Vikram Pudi. 2021.
\newblock \href {https://doi.org/10.18653/v1/2021.findings-emnlp.230}
  {Adversarial examples for evaluating math word problem solvers}.
\newblock In \emph{Findings of the Association for Computational Linguistics:
  {EMNLP} 2021, Virtual Event / Punta Cana, Dominican Republic, 16-20 November,
  2021}, pages 2705--2712. Association for Computational Linguistics.

\bibitem[{Kumar et~al.(2022)Kumar, Maheshwary, and Pudi}]{kumar2022practice}
Vivek Kumar, Rishabh Maheshwary, and Vikram Pudi. 2022.
\newblock \href {https://doi.org/10.18653/v1/2022.naacl-main.310} {Practice
  makes a solver perfect: Data augmentation for math word problem solvers}.
\newblock In \emph{Proceedings of the 2022 Conference of the North American
  Chapter of the Association for Computational Linguistics: Human Language
  Technologies, {NAACL} 2022, Seattle, WA, United States, July 10-15, 2022},
  pages 4194--4206. Association for Computational Linguistics.

\bibitem[{Le~Bras et~al.(2020)Le~Bras, Swayamdipta, Bhagavatula, Zellers,
  Peters, Sabharwal, and Choi}]{le2020adversarial}
Ronan Le~Bras, Swabha Swayamdipta, Chandra Bhagavatula, Rowan Zellers, Matthew
  Peters, Ashish Sabharwal, and Yejin Choi. 2020.
\newblock Adversarial filters of dataset biases.
\newblock In \emph{International Conference on Machine Learning}, pages
  1078--1088. PMLR.

\bibitem[{Li et~al.(2022{\natexlab{a}})Li, Xiao, Liang, and
  Chen}]{Li2022semantic}
Ailisi Li, Yanghua Xiao, Jiaqing Liang, and Yunwen Chen. 2022{\natexlab{a}}.
\newblock Semantic-based data augmentation for math word problems.
\newblock In \emph{International Conference on Database Systems for Advanced
  Applications}, pages 36--51. Springer.

\bibitem[{Li et~al.(2019)Li, Gao, Bing, King, and Lyu}]{li2019improving}
Jingjing Li, Yifan Gao, Lidong Bing, Irwin King, and Michael~R. Lyu. 2019.
\newblock \href {https://doi.org/10.18653/v1/D19-1317} {Improving question
  generation with to the point context}.
\newblock In \emph{Proceedings of the 2019 Conference on Empirical Methods in
  Natural Language Processing and the 9th International Joint Conference on
  Natural Language Processing, {EMNLP-IJCNLP} 2019, Hong Kong, China, November
  3-7, 2019}, pages 3214--3224. Association for Computational Linguistics.

\bibitem[{Li et~al.(2022{\natexlab{b}})Li, Zhang, Yan, Zhou, Li, Liu, and
  Cao}]{li2021seeking}
Zhongli Li, Wenxuan Zhang, Chao Yan, Qingyu Zhou, Chao Li, Hongzhi Liu, and
  Yunbo Cao. 2022{\natexlab{b}}.
\newblock \href {https://doi.org/10.18653/v1/2022.findings-acl.195} {Seeking
  patterns, not just memorizing procedures: Contrastive learning for solving
  math word problems}.
\newblock In \emph{Findings of the Association for Computational Linguistics:
  {ACL} 2022, Dublin, Ireland, May 22-27, 2022}, pages 2486--2496. Association
  for Computational Linguistics.

\bibitem[{Liang et~al.(2022)Liang, Zhang, Wang, Qin, Lan, Shao, and
  Zhang}]{liang2022mwp}
Zhenwen Liang, Jipeng Zhang, Lei Wang, Wei Qin, Yunshi Lan, Jie Shao, and
  Xiangliang Zhang. 2022.
\newblock Mwp-bert: Numeracy-augmented pre-training for math word problem
  solving.
\newblock In \emph{Findings of the Association for Computational Linguistics:
  NAACL 2022}, pages 997--1009.

\bibitem[{Lin(2004)}]{lin2004rouge}
Chin-Yew Lin. 2004.
\newblock Rouge: A package for automatic evaluation of summaries.
\newblock In \emph{Text summarization branches out}, pages 74--81.

\bibitem[{Liu et~al.(2021{\natexlab{a}})Liu, Guan, Li, Cheng, Kawahara, and
  Kurohashi}]{liu2021roda}
Qianying Liu, Wenyu Guan, Sujian Li, Fei Cheng, Daisuke Kawahara, and Sadao
  Kurohashi. 2021{\natexlab{a}}.
\newblock Roda: Reverse operation based data augmentation for solving math word
  problems.
\newblock \emph{IEEE/ACM Transactions on Audio, Speech, and Language
  Processing}, 30:1--11.

\bibitem[{Liu et~al.(2021{\natexlab{b}})Liu, Fang, Ding, Li, Wu, and
  Liu}]{liu2021mathematical}
Tianqiao Liu, Qiang Fang, Wenbiao Ding, Hang Li, Zhongqin Wu, and Zitao Liu.
  2021{\natexlab{b}}.
\newblock Mathematical word problem generation from commonsense knowledge graph
  and equations.
\newblock In \emph{Proceedings of the 2021 Conference on Empirical Methods in
  Natural Language Processing}, pages 4225--4240.

\bibitem[{Liyanage and Ranathunga(2020)}]{liyanage2020multi}
Vijini Liyanage and Surangika Ranathunga. 2020.
\newblock Multi-lingual mathematical word problem generation using long short
  term memory networks with enhanced input features.
\newblock In \emph{Proceedings of The 12th Language Resources and Evaluation
  Conference}, pages 4709--4716.

\bibitem[{Miao et~al.(2020)Miao, Liang, and Su}]{miao2020diverse}
Shen-Yun Miao, Chao-Chun Liang, and Keh-Yih Su. 2020.
\newblock A diverse corpus for evaluating and developing english math word
  problem solvers.
\newblock In \emph{Proceedings of the 58th Annual Meeting of the Association
  for Computational Linguistics}, pages 975--984.

\bibitem[{Moon-Rembert and Gilbert(2019)}]{moon2019illmatics}
DeKita~G Moon-Rembert and Juan~E Gilbert. 2019.
\newblock Illmatics: A web-based math word problem generator for students’
  distal and proximal interests.
\newblock In \emph{E-Learn: World Conference on E-Learning in Corporate,
  Government, Healthcare, and Higher Education}, pages 842--848. Association
  for the Advancement of Computing in Education (AACE).

\bibitem[{Ou et~al.(2022)Ou, Zhang, Feng, and Zhou}]{ou2022counterfactual}
Jiao Ou, Jinchao Zhang, Yang Feng, and Jie Zhou. 2022.
\newblock \href {https://aclanthology.org/2022.emnlp-main.106} {Counterfactual
  data augmentation via perspective transition for open-domain dialogues}.
\newblock In \emph{Proceedings of the 2022 Conference on Empirical Methods in
  Natural Language Processing, {EMNLP} 2022, Abu Dhabi, United Arab Emirates,
  December 7-11, 2022}, pages 1635--1648. Association for Computational
  Linguistics.

\bibitem[{Papineni et~al.(2002)Papineni, Roukos, Ward, and
  Zhu}]{papineni2002bleu}
Kishore Papineni, Salim Roukos, Todd Ward, and Wei-Jing Zhu. 2002.
\newblock Bleu: a method for automatic evaluation of machine translation.
\newblock In \emph{Proceedings of the 40th annual meeting of the Association
  for Computational Linguistics}, pages 311--318.

\bibitem[{Patel et~al.(2021)Patel, Bhattamishra, and Goyal}]{patel2021nlp}
Arkil Patel, Satwik Bhattamishra, and Navin Goyal. 2021.
\newblock \href {https://doi.org/10.18653/v1/2021.naacl-main.168} {Are {NLP}
  models really able to solve simple math word problems?}
\newblock In \emph{Proceedings of the 2021 Conference of the North American
  Chapter of the Association for Computational Linguistics: Human Language
  Technologies, {NAACL-HLT} 2021, Online, June 6-11, 2021}, pages 2080--2094.
  Association for Computational Linguistics.

\bibitem[{Polozov et~al.(2015)Polozov, O'Rourke, Smith, Zettlemoyer, Gulwani,
  and Popovi{\'c}}]{polozov2015personalized}
Oleksandr Polozov, Eleanor O'Rourke, Adam~M Smith, Luke Zettlemoyer, Sumit
  Gulwani, and Zoran Popovi{\'c}. 2015.
\newblock Personalized mathematical word problem generation.
\newblock In \emph{Twenty-Fourth International Joint Conference on Artificial
  Intelligence}.

\bibitem[{Rothe et~al.(2020)Rothe, Narayan, and Severyn}]{rothe2020leveraging}
Sascha Rothe, Shashi Narayan, and Aliaksei Severyn. 2020.
\newblock Leveraging pre-trained checkpoints for sequence generation tasks.
\newblock \emph{Transactions of the Association for Computational Linguistics},
  8:264--280.

\bibitem[{Schick and Sch{\"{u}}tze(2021)}]{schick2020exploiting}
Timo Schick and Hinrich Sch{\"{u}}tze. 2021.
\newblock \href {https://doi.org/10.18653/v1/2021.eacl-main.20} {Exploiting
  cloze-questions for few-shot text classification and natural language
  inference}.
\newblock In \emph{Proceedings of the 16th Conference of the European Chapter
  of the Association for Computational Linguistics: Main Volume, {EACL} 2021,
  Online, April 19 - 23, 2021}, pages 255--269. Association for Computational
  Linguistics.

\bibitem[{Shridhar et~al.(2022)Shridhar, Macina, El{-}Assady, Sinha, Kapur, and
  Sachan}]{shridhar2022automatic}
Kumar Shridhar, Jakub Macina, Mennatallah El{-}Assady, Tanmay Sinha, Manu
  Kapur, and Mrinmaya Sachan. 2022.
\newblock \href {https://aclanthology.org/2022.emnlp-main.277} {Automatic
  generation of socratic subquestions for teaching math word problems}.
\newblock In \emph{Proceedings of the 2022 Conference on Empirical Methods in
  Natural Language Processing, {EMNLP} 2022, Abu Dhabi, United Arab Emirates,
  December 7-11, 2022}, pages 4136--4149. Association for Computational
  Linguistics.

\bibitem[{Sun et~al.(2018)Sun, Liu, Lyu, He, Ma, and Wang}]{sun2018answer}
Xingwu Sun, Jing Liu, Yajuan Lyu, Wei He, Yanjun Ma, and Shi Wang. 2018.
\newblock Answer-focused and position-aware neural question generation.
\newblock In \emph{Proceedings of the 2018 Conference on Empirical Methods in
  Natural Language Processing}, pages 3930--3939.

\bibitem[{Sundaram et~al.(2022)Sundaram, Gurajada, Fisichella, Abraham
  et~al.}]{sundaram2022nlp}
Sowmya~S Sundaram, Sairam Gurajada, Marco Fisichella, Savitha~Sam Abraham,
  et~al. 2022.
\newblock Why are nlp models fumbling at elementary math? a survey of deep
  learning based word problem solvers.
\newblock \emph{arXiv preprint arXiv:2205.15683}.

\bibitem[{Tai et~al.(2015)Tai, Socher, and Manning}]{tai2015improved}
Kai~Sheng Tai, Richard Socher, and Christopher~D. Manning. 2015.
\newblock \href {https://doi.org/10.3115/v1/p15-1150} {Improved semantic
  representations from tree-structured long short-term memory networks}.
\newblock In \emph{Proceedings of the 53rd Annual Meeting of the Association
  for Computational Linguistics and the 7th International Joint Conference on
  Natural Language Processing of the Asian Federation of Natural Language
  Processing, {ACL} 2015, July 26-31, 2015, Beijing, China, Volume 1: Long
  Papers}, pages 1556--1566. The Association for Computer Linguistics.

\bibitem[{Tan et~al.(2021)Tan, Wang, Jiang, and Jiang}]{tan2021investigating}
Minghuan Tan, Lei Wang, Lingxiao Jiang, and Jing Jiang. 2021.
\newblock Investigating math word problems using pretrained multilingual
  language models.
\newblock \emph{arXiv preprint arXiv:2105.08928}.

\bibitem[{Wang et~al.(2017)Wang, Liu, and Shi}]{wang2017deep}
Yan Wang, Xiaojiang Liu, and Shuming Shi. 2017.
\newblock Deep neural solver for math word problems.
\newblock In \emph{Proceedings of the 2017 Conference on Empirical Methods in
  Natural Language Processing}, pages 845--854.

\bibitem[{Wang et~al.(2021)Wang, Lan, and Baraniuk}]{wang2021math}
Zichao Wang, Andrew~S. Lan, and Richard~G. Baraniuk. 2021.
\newblock \href {https://doi.org/10.18653/v1/2021.emnlp-main.484} {Math word
  problem generation with mathematical consistency and problem context
  constraints}.
\newblock In \emph{Proceedings of the 2021 Conference on Empirical Methods in
  Natural Language Processing, {EMNLP} 2021, Virtual Event / Punta Cana,
  Dominican Republic, 7-11 November, 2021}, pages 5986--5999. Association for
  Computational Linguistics.

\bibitem[{Williams(2011)}]{williams2011generating}
Sandra Williams. 2011.
\newblock Generating mathematical word problems.
\newblock In \emph{2011 AAAI Fall symposium series}.

\bibitem[{Xie et~al.(2021)Xie, Lv, Xia, Wu, Qin, Liu, and Yan}]{xie2021target}
Shufang Xie, Ang Lv, Yingce Xia, Lijun Wu, Tao Qin, Tie-Yan Liu, and Rui Yan.
  2021.
\newblock Target-side input augmentation for sequence to sequence generation.
\newblock In \emph{International Conference on Learning Representations}.

\bibitem[{Xie and Sun(2019)}]{xie2019goal}
Zhipeng Xie and Shichao Sun. 2019.
\newblock A goal-driven tree-structured neural model for math word problems.
\newblock In \emph{IJCAI}, pages 5299--5305.

\bibitem[{Yang et~al.(2022)Yang, Qin, Chen, and Liang}]{yang2022unbiased}
Zhicheng Yang, Jinghui Qin, Jiaqi Chen, and Xiaodan Liang. 2022.
\newblock \href {https://doi.org/10.18653/v1/2022.findings-naacl.104} {Unbiased
  math word problems benchmark for mitigating solving bias}.
\newblock In \emph{Findings of the Association for Computational Linguistics:
  {NAACL} 2022, Seattle, WA, United States, July 10-15, 2022}, pages
  1401--1408. Association for Computational Linguistics.

\bibitem[{Zhang et~al.(2019)Zhang, Wang, Zhang, Dai, and Shen}]{zhang2019gap}
Dongxiang Zhang, Lei Wang, Luming Zhang, Bing~Tian Dai, and Heng~Tao Shen.
  2019.
\newblock The gap of semantic parsing: A survey on automatic math word problem
  solvers.
\newblock \emph{IEEE transactions on pattern analysis and machine
  intelligence}, 42(9):2287--2305.

\bibitem[{Zhang et~al.(2020{\natexlab{a}})Zhang, Wang, Lee, Bin, Wang, Shao,
  and Lim}]{zhang2020graph}
Jipeng Zhang, Lei Wang, Roy Ka-Wei Lee, Yi~Bin, Yan Wang, Jie Shao, and Ee-Peng
  Lim. 2020{\natexlab{a}}.
\newblock Graph-to-tree learning for solving math word problems.
\newblock Association for Computational Linguistics.

\bibitem[{Zhang et~al.(2020{\natexlab{b}})Zhang, Kishore, Wu, Weinberger, and
  Artzi}]{zhang2019bertscore}
Tianyi Zhang, Varsha Kishore, Felix Wu, Kilian~Q. Weinberger, and Yoav Artzi.
  2020{\natexlab{b}}.
\newblock \href {https://openreview.net/forum?id=SkeHuCVFDr} {Bertscore:
  Evaluating text generation with {BERT}}.
\newblock In \emph{8th International Conference on Learning Representations,
  {ICLR} 2020, Addis Ababa, Ethiopia, April 26-30, 2020}. OpenReview.net.

\end{thebibliography}
\bibliographystyle{acl_natbib}
\appendix
\section{Generated data by DQGF}
Table~\ref{tab_app} shows five examples of the generated data by DQGF. Original data is the MWP in dataset which only has single question for each scenario. Generated data is the diverse questions with equations on original data generated by our DQGF.
\label{sec:appendix}
\begin{table*}[t]
\centering
\scalebox{0.68}{
\begin{tabular}{|l|}
\hline
\begin{tabular}[c]{@{}l@{}}\textcolor{blue}{Original Data}\\ Text: A pair of pants costs 58 dollars, and a jacket costs 4 times as much as a pair of pants. \\ How many dollars are spent on 5 sets of these clothes?\\ Equation: x=5*(58+(58*4))\end{tabular} \\\\
\begin{tabular}[c]{@{}l@{}}\textcolor{red}{Generated Data}\\ Question: How much do a pair of pants and a jacket cost in total?\\ Equation:x=58+58*4\\ \\ Question: How much does a jacket cost?\\ Equation: x=58*4\end{tabular} \\ \hline
\begin{tabular}[c]{@{}l@{}}\textcolor{blue}{Original Data}\\ Text: Dingding has read 180 pages of a book and has 150 pages left to read. \\ How many pages are there in this book?\\ Equation: x=180+150\end{tabular} \\\\
\begin{tabular}[c]{@{}l@{}}\textcolor{red}{Generated Data}\\ Question: How many more pages have been read than have not been read?\\ Equation: x=180-150\\ \\ Question: How many times more pages have been read than have not been read?\\ Equation: x=180/150\end{tabular} \\ \hline
\begin{tabular}[c]{@{}l@{}}\textcolor{blue}{Original Data}\\ Text: Qiangqiang's father and mother work outside. Father sends Qiangqiang 458 dollars a month\\ and mother sends Qiangqiang 447 dollars a month. How much money do Qiangqiang's father and \\ mother send to Qiangqiang each month?\\ Equation: x=458+447\end{tabular} \\\\
\begin{tabular}[c]{@{}l@{}}\textcolor{red}{Generated Data}\\ Question: How much more money does the mother send to Qiangqiang each month \\ than the father?\\ Equation: x=447-458\\ \\ Question: How many times more money does the mother send to Qiangqiang than \\ the father each month?\\ x=447/458\\ \\ Question: How much more money does the father send to Qiangqiang each month \\ than the mother?\\ Equation: x=458-447\\ \\ Question: How many times more money does the father send to Qiangqiang each \\ month than the mother?\\ Equation: x=458/447\end{tabular} \\ \hline
\begin{tabular}[c]{@{}l@{}}\textcolor{blue}{Original Data}\\ Text: Mom bought a toothbrush for 3.6 dollars and a box of toothpaste for 9.5 dollars. How much \\ is a toothbrush cheaper than a box of toothpaste?\\ Equation: x=9.5-3.6\end{tabular} \\\\
\begin{tabular}[c]{@{}l@{}}\textcolor{red}{Generated Data}\\ Question: What is the ratio of the price of a box of toothpaste to a toothbrush?\\ Equation: x=9.5/3.6\\ \\ Question: How much do a toothbrush and a box of toothpaste cost in total?\\ Equation: x=3.6+9.5\\ \\ Question: How much more expensive is a toothbrush than a box of toothpaste?\\ Equation: x=3.6-9.5\\ \\ What is the ratio of the price of a toothbrush to a box of toothpaste?\\ Equation: x=3.6/9.5\end{tabular} \\ \hline
\begin{tabular}[c]{@{}l@{}}\textcolor{blue}{Original Data}\\ Text: A storybook has 438 pages and Xiao Liang has read 202 pages. How many pages does Xiao Liang \\ have left to read?\\ Equation: x=438-202\end{tabular} \\\\
\begin{tabular}[c]{@{}l@{}}\textcolor{red}{Generated Data}\\ Question: What is the ratio of the number of pages Xiao Liang has read to the total \\ number of pages in the storybook?\\ Equation: x=202/438\\ \\ Question: How many times is the total number of pages in the storybook than the number \\ of pages Xiao Liang has read?\\ Equation: x=438/202\end{tabular} \\ \hline
\end{tabular}
}
\caption{Five generated MWP samples with \textcolor{blue}{Original data} and \textcolor{red}{Generated diverse questions with equations} by DQGF}
\label{tab_app}
\end{table*}

\end{document}